# Benchmark Performance of Machine And Deep Learning Based Methodologies for Urdu Text Document Classification

**Muhammad Nabeel Asim[1], Muhammad Usman Ghani[2], Muhammad Ali Ibrahim[3], Sheraz Ahmad[4], Waqar Mahmood[5], Andreas Dengel[6]**

**Abstract**

In order to provide benchmark performance for Urdu text document classification, the contribution of this paper is manifold. First, it pro-vides a publicly available benchmark dataset manually tagged against 6 classes. Second, it investigates the performance impact of traditional machine learning based Urdu text document classification methodologies by embedding 10 filter-based feature selection algorithms which have been widely used for other languages. Third, for the very first time, it as-sesses the performance of various deep learning based methodologies for Urdu text document classification. In this regard, for experimentation, we adapt 10 deep learning classification methodologies which have pro-duced best performance figures for English text classification. Fourth, it also investigates the performance impact of transfer learning by utiliz-ing Bidirectional Encoder Representations from Transformers approach for Urdu language. Fifth, it evaluates the integrity of a hybrid approach which combines traditional machine learning based feature engineering and deep learning based automated feature engineering. Experimental results show that feature selection approach named as Normalised Dif-ference Measure along with Support Vector Machine outshines state-of-the-art performance on two closed source benchmark datasets CLE Urdu Digest $1000k$, and CLE Urdu Digest $1Million$ with a significant margin of 32%, and 13% respectively. Across all three datasets, Normalised Differ-ence Measure outperforms other filter based feature selection algorithms as it significantly uplifts the performance of all adopted machine learning, deep learning, and hybrid approaches. The source code and presented dataset are available at Github repository [1].

[1]https://github.com/minixain/Urdu-Text-Classification

# 1 Introduction

Textual resources of diverse domains such as academia and industries are growing enormously over the web due to the rapid growth of technology [1, 2]. According to a recent survey of data facts, users have only utilized $0.5\%$ of all electronic textual data[3]. The amount of electronic textual data which has been created in last two years is way more than the data created by entire human race previously[3]. This marks the desperate need of classifying or categorizing such humongous electronic textual data in order to enable the processing of text at large scale and for the extraction of useful insights. With the emergence of computational methodologies for text classification, multifarious applications have been developed such as Email Spam detection[4], Gender identification[5], Product review analysis[6], News categorization[7, 8, 9] and Fake news detection [10, 11, 12] for various languages like English, Arabic, and Chinese. However, despite crossing the landmark of 100 $million$ speakers [13], Urdu language is still lacking in the development of such applications. The primary reason behind this limited progress is the lack of publicly available datasets for Urdu language. Urdu text document classification datasets used in the previous works are pri-vate [14], [15], [16], [17], [18], [19] which further restricts the research and fair comparison of new methodologies. In order to overcome this limitation, the pa-per in hand provides a new publicly available dataset in which news documents are manually tagged against six different classes.

On the other hand, regarding the improvement in performance of traditional machine learning based text document classification methodologies, feature selection has played a significant role in various languages such as English, Arabic, and Chinese [20], [21]. The ultimate aim of feature selection is to eliminate irrelevant and redundant features [22]. Feature selection alleviates the burden on classifier which leads to faster training [23], [24]. It also assists the classifier to draw better decision boundary which eventually results in accurate predictions [23], [24]. State-of-the-art machine learning based Urdu text document classification methodologies lack discriminative feature selection techniques [19]. In this paper, we embed ten most anticipated filter based feature ranking metrics in traditional machine learning pipeline to extrapolate the impact created by the set of selected top $k$ features over the performance of Support Vector Machine (SVM) [25] and Naive Bayes (NB) [26] classifiers.

Although feature selection techniques reduce the dimensionality of textual data up to great extent, however, traditional machine learning based text document classification methodologies still face the sparsity problem in bag of words based feature representation techniques [27], [28]. Bag of words based feature representation techniques consider unigrams, n-grams or specific patterns as features [27], [28]. These algorithms do not capture the complete contextual information of data and also face the problem of data sparsity [27], [28]. These problems are solved by word embeddings which do not only capture syntactic but semantic information of textual data as well [29]. Deep learning based text document classification methodologies provide end to end system for text classification by automating the process of feature engineering and are outperforming state-of-the-art machine learning based classification approaches [30] [31].

Although there exists some work on the development of pre-trained neural word embeddings (Haider et. al [32], and FastText [2]) for Urdu language, however, no researcher has utilized any deep learning based methodology or pre-trained neural word embeddings for Urdu text document classification. Here we thoroughly investigate the performance impact of 10 state-of-the-art deep learn-ing methodologies using pre-trained neural word embeddings. Amongst all, 4 methodologies are based on a convolutional neural network (CNN), 3 on a recurrent neural network (RNN), and 3 of them are based on a hybrid approach (CNN+RNN). Pre-trained neural word embeddings are just shallow representations as they fuse learned knowledge only in the very first layer of deep learning model. Whereas, rest of the layers still require to be trained using randomly initialized weights of various filters [33]. Moreover, although pre-trained neural word embeddings manage to capture semantic information of words but fail to acquire high level information including long range dependencies, anaphora, negation, and agreement for different domains [34], [35], [33]. Considering the recent trend of utilizing pre-trained language models to overcome the downfalls of pre-trained neural word embeddings [36],[37], we also explore the impact of language modelling for the task of Urdu text document classification.

However, due to the lack of extensive research, finding an optimal way to acquire maximal results on diverse natural language processing tasks through the use of Bidirectional Encoder Representations from Transformers (BERT) [38] is not straightforward at all [39], [40], [41]. For instance, whether pre-training BERT [38] on domain-specific data will produce good results, or fine-tuning BERT [38] for target tasks or multitask learning would be an optimal option [39], [40], [41]. In this paper, we thoroughly investigate multifarious ways to fine-tune pre-trained multilingual BERT [38] language models and provide key insights to make the best use of BERT [38] for Urdu text document classification.

Previously, we proposed a robust machine and deep learning based hybrid approach [42] for English text document classification. The proposed hybrid methodology reaped the benefits of both machine learning based feature engineering and automated engineering performed by deep learning models which eventually helped the model to better classify text documents into predefined classes [42]. Hybrid approach significantly improved the performance of text document classification on two publicly available benchmark English datasets 20-Newsgroup [3], and BBC [4] [42]. This paper investigates whether utilization of both machine and deep learning based feature engineering is versatile and ef-fective enough to replicate promising performance figures with a variety of deep learning models for Urdu text document classification. Extensive experimenta-tion with all machine and deep learning based methodologies is performed on two closed source datasets namely CLE Urdu Digest $1000k$, CLE Urdu Digest $1Million$, and one newly developed dataset namely DSL Urdu news.

[2] https://fasttext.cc/docs/en/crawl-vectors.html

[3] http://archive.ics.uci.edu/ml/datasets/twenty+newsgroups

[4] http://mlg.ucd.ie/datasets/bbc.html

Amongst all machine learning based methodologies, Naive Bayes [26] with Normalized Difference Measure [43] marks the highest performance of 94% over newly developed DSL Urdu News dataset. Whereas, SVM [25] proves domi-nant over both close source datasets CLE Urdu Digest $1000k$, CLE Urdu Digest $1Million$ by marking the performance of 92% with Normalized Difference Measure [43], and 83% with Chi-Squared (CHISQ) [44]. On the other hand, trivial adopted deep learning based methodologies manage to outshine state-of-the-art performance by the margin of 6% on CLE Urdu Digest $1000k$, and 1% on CLE Urdu Digest $1Million$. Contrarily, hybrid methodology which leverages machine and deep learning based feature engineering [42], and BERT [38] mark similar performance across all three datasets. These methodologies outperform state-of-the-art performance with the figure of 18% on CLE Urdu Digest $1000k$, 10% on CLE Urdu Digest $1M$ datasets, and almost equalize the promising performance figures of machine learning based methodology over DSL Urdu news dataset. Primary contributions of this paper can be summarized as:

1. Development of a publicly available dataset that contains 662 documents of six different classes (health-science, sports, business, agriculture, world, and entertainment) containing 130 $K$ words for Urdu text document clas-sification.
2. Benchmarking performance of Urdu text document classification by employing 10 filter based feature selection algorithms such as Balanced Accu-racy Measure (ACC2) [45], Normalized Difference Measure (NDM) [43], Max-Min Ratio (MMR) [21], Relative Discrimination Criterion (RDC) [46], Information Gain (IG) [47], Chi-Squared (CHISQ) [44], Odds Ra-tio (OR) [48], Bi-Normal Separation (BNS) [45], Gini Index (GINI) [49], Poisson Ratio (POISON) [50] in traditional machine learning pipeline. Furthermore, making comparison among feature selection algorithms on predefined benchmark test points in order to determine which algorithm selects optimal set of features that can further be used to develop various applications for Urdu language.
3. In order to compare the performance of machine and deep learning based methodologies, we adapt 10 state-of-the-art deep learning based text doc-ument classification methodologies.
4. Considering the little research to optimize Bidirectional Encoder Representations from Transformers (BERT [38]) for acquiring better performance over target tasks, we explore multifarious ways to optimally fine-tune pre-trained multilingual BERT [38] to provide benchmark performance for the task of Urdu text document classification. In addition, we facilitate key observations and a generic solution to fine-tune BERT [38] for text classification.
5. The fruitfulness of hybrid approach which reaps the benefits of traditional feature engineering and deep learning based automated feature engineering is thoroughly investigated with a variety of deep learning models and Urdu datasets.

The remaining paper is distributed into following sections. First section discusses previous work solely related to Urdu Text Document Classification followed by a detail explanation of text document classification methodologies used in this paper. Then, all datasets are elaborated comprehensively. Afterwards, experimental setup and results are revealed in subsequent sections. Finally, we summarize the key points and give future directions.

## 2 Related Work

Text document classification methodologies can be categorized into rule-based and statistical approaches. Rule-based approaches utilize manually written lin-guistic rules, whereas, statistical approaches learn the association among multi-farious features and class labels in order to classify text documents into prede-fined classes [51]. This section briefly illustrates state-of-the-art statistical work on Urdu text document classification.

Ali et al. [14] compared the performance of two classifiers namely Naïve Bayes (NB) [26], and Support Vector Machine (SVM) [25] for the task of Urdu text document classification. They prepared a dataset by scrapping various Urdu news websites and manually classified them into six categories (news, sports, finance, culture, consumer information and personal information). Based on their experimental results, they summarized that SVM [25] significantly out performed Naive Bayes [26]. Their experiments also revealed that stemming decreased the overall performance of classification.

Usman et al. [15] utilized maximum voting approach in quest of classifying Urdu News documents. The news corpus was divided into seven categories namely business, entertainment, culture, health, sports, and weird. After tokenization, stop words removal, and stemming, they extracted 93400 terms and fed them to six machine learning classifiers namely Naïve Bayes, Linear Stochastic Gradient Descent (SGD)[52], Multinomial Naïve Bayes, Bernoulli Naïve Bayes [53], Linear SVM [25], and Random Forest Classifier[54]. Then, they applied max voting approach in such a way that the class selected by majority of the classification algorithms was chosen as final class. Experimen-tally they proved that, Linear SVM [25] and Linear SGD [52] showed better performance on their developed corpora.

Sattar et al.[17] performed Urdu editorials classification using Naïve Bayes classifier. Moreover, most frequent terms of the corpus were removed to alleviate the dimensionality of data. Their experimental results showed that Naïve Bayes classifier performs well when it is fed with frequent terms as compared to feeding all unique terms of the corpus.

Ahmed et al. [16] performed Urdu news headlines classification using support vector machine (SVM) [25] classifier. They utilized a TF-IDF based feature selection approach which removed less important domain specific terms from underly corpus. This was done by utilizing the threshold paradigm on TF-IDF score which enabled the extraction of those terms that had higher TFIDF than defined threshold value. After preprocessing and threshold based term filtration, they used SVM [25] classifier to make predictions.

Zia et al. [18] evaluated the performance of Urdu text document classification by adopting four state-of-the-art feature selection techniques namely Informa-tion Gain (IG) [47], Chi Square (CS)[44], Gain Ratio (GR)[55], and Symmet-rical Uncertainty[56] with four classification algorithms (K-Nearest Neighbors (KNN)[57], Naïve Bayes (NB), Decision Tree (DT), and Support Vector Ma-chines (SVM) [25]. They found that for larger datasets, performance of SVM [25] with any of the above mentioned feature selection technique was better as compared to Naïve Bayes [26] which was more inclined towards small corpora.

Adeeba et al.[19] presented an automatic Urdu text based genre identifica-tion system that classified Urdu text documents into one of the eight predefined categories namely culture, science, religion, press, health, sports, letters and interviews. They investigated the effects of employing both lexical and struc-tural features on the performance of Support Vector Machine [25], Naïve Bayes [26] and Decision Tree algorithms. For lexical features, the authors extracted word unigrams, bigrams, along with their term frequency and inverse document frequency. To extract structural features, part of speech tags and word sense information were utilized. Moreover, they reduced the dimensionality of corpora by eliminating low frequency terms. For the experimentation, CLE Urdu Digest 100K[5] and CLE Urdu Digest 1 Million[6] corpora were used. Their experiments revealed that SVM [25] was better than other classifiers irrespective of feature types.

State-of-the-art work on Urdu text document classification is summarized in the Table 1 by author name, benchmark dataset, exploited feature representa-tion and selection techniques, classifiers, evaluation metrics, and their respective performances.

| Authors | Datasets | Feature Representation Techniques | Feature Selection Techniques | Classifier | Evaluation Metric |
|---|---|---|---|---|---|
| Ali et al [14] | Manually Classified News Corpus | Normalized Term Frequency | — | NB , SVM | Accuracy |
| Usman et al [15] | News Copus | Term Frequency (TF) | — | NB, BNB, LSVM, LSGB, RF | Precision, Recall, F1-score |
| Sattar et al [17] | Urdu News Editorials | Term Frequency (TF) | — | NB | Precision, Recall, F1-score |
| Ahmed et al [16] | Urdu News Headlines | TF-IDF | TF-IDF (Thresholding) | SVM | Accuracy |
| Zia et al [18] | EMILLE, Self Collected Naive corpus (News) | TF-IDF | Information Gain, Chi Square, Gain Ratio, Symmetrical Uncertainty | KNN, DT, NB. | F1-score |
| Adeeba et al [19] | CLE Urdu Digest ($1000K$, $1Million$) | Term Frequency (TF), TF-IDF | Pruning | SVM (Linear, Polynomial, Radial), NB | Precision, Recall, F1-score |

Table 1: State-of-the-art work on Urdu Text document classification

After thoroughly examining the literature, it can be summarized that SVM [25] and Naive Bayes [26] perform better than other classifiers for the task of Urdu text document classification.

[5]http://www.cle.org.pk/clestore/urdudigestcorpus100k.htm

[6]http://www.cle.org.pk/clestore/urdudigestcorpus1M.htm

For English text document classification, recent experimentation on public benchmark datasets also proves that performance of SVM [25], and Naive Bayes [26] significantly improves with the use of filter based feature selection algorithms [21], [43]. Filter based feature selection algorithms do not only improve the performance of machine learning based methodologies but it has also substantially raised the performance of deep learning based text document clas-sification approaches [42].

However, Urdu text document classification methodologies are lacking to produce promising performance due to the lack of research in this direction as only Ahmed et al [16], and Zia et al [18] utilized some feature selection approaches in order to reduce the dimensionality of data. While Ahmed et al [16] only experimented with TF-IDF based feature selection approach, Zia et al. [18] assessed the integrity of just four feature selection algorithms (Information Gain (IG) [47], Chi Square (CS), Gain Ratio (GR), and Symmetrical Uncertainty) in domain of Urdu text document classification. However, the performance impact of more recent filter based feature selection algorithms has never been explored specifically for Urdu text document classification.

In addition, despite the promising performance produced by deep learning methodologies for diverse NLP tasks [58], [59], no researcher has utilized any deep learning based methodology for the task of Urdu text document classifica-tion.

# 3 Adopted Methodologies For Urdu Text Document Classification

This section comprehensively illustrates machine learning, deep learning, and hybrid methodologies which we have used for the task of Urdu text document classification.

## 3.1 Traditional Machine Learning Based Urdu Text Document Classification With Filter Based Feature Selection Algorithm

This section elaborates the machine learning based Urdu text document classification methodology. Primarily, our main focus is to investigate the performance boost in traditional machine learning based Urdu text document classi-fication methodologies through the embedding of filter based feature selection algorithms. Figure 1 provides graphical illustration of machine learning based Urdu text classification methodology which utilizes filter based feature engineer-ing. All phases of this methodology are discussed below.

## 3.2 Preprocessing

Preprocessing of text is considered as preliminary step in almost all natural language processing tasks as better tokenization, and stemming or lemmatization eventually leads to better performance in various machine learning tasks such as Text Classification [60], [61], Information Retrieval [62] and Text Summarization [63].

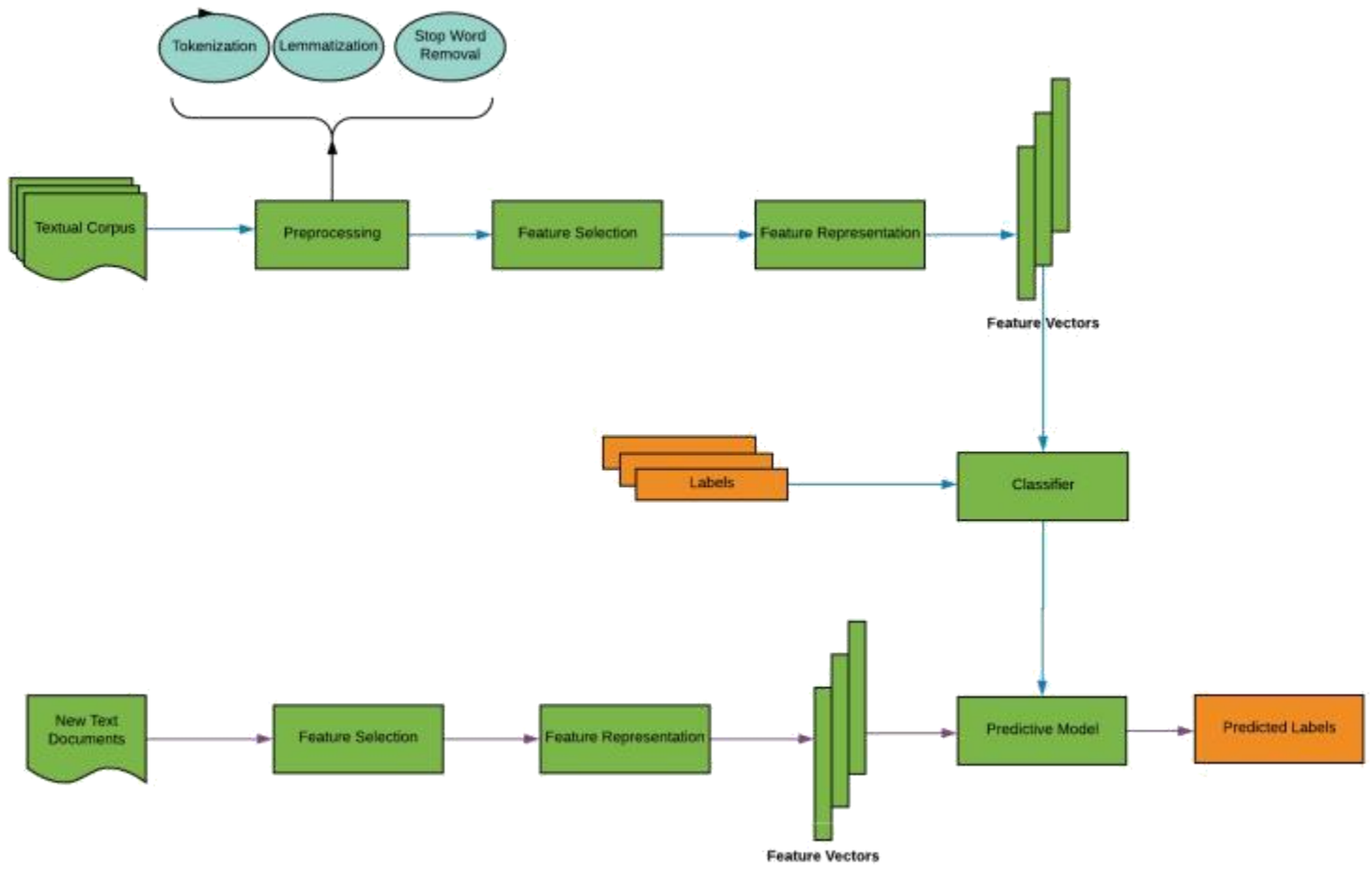

Figure 1: Machine Learning Based Urdu Text Document Classification Method-ology

Stemming undoubtedly plays an important role to alleviate sparsity prob-lems through dimensionality reduction, however, there are very few rule based stemmers available for Urdu language which lack to showcase quality perfor-mance. Ali et al. [14] claimed that stemming degrades the performance of Urdu text document classification. We analyzed that the stemmer utilized by Ali et al [14] was of poor quality which eventually caused the decline in performance as it has been proved by many researchers that stemming often improves the performance of text document classification for various languages (e.g English) [64], [65]. Urdu language lacks better stemming algorithms, therefore, instead of stemming, we perform lemmatization through a manually prepared Urdu lex-icon containing $9743$ possible values of $4162$ base terms. In section 4, Tables 2, 3, 4 reveal the impact of lemmatization on the size reduction of three datasets used in our experimentation. We believe public access to the developed lexicon will enable the researchers to perform lemmatization in several different Urdu processing tasks. In addition, all non significant words of corpus are eliminated through a stop words list. The list of $1000$ stop words is formed by manually analyzing the most frequent $1500$ words of underlay corpora.

## 3.3 Feature Selection

Feature selection is being widely used to reduce the dimensionality of feature space in different applications like text classification [43], plagiarism detection [66], and for query expansion in pseudo relevance feedback based information retrieval [67], which eventually assists to produce better results.

Feature selection approaches can be categorized into three classes wrapper [68], embedded [69], and filter [70]. In wrapper methods, classifier is trained and tested over several subsets of features and only one subset of features is selected which has produced the minimum error [68]. Similarly, embedded feature selection approaches also work like wrapper based methods, however wrapper based methods can exploit one classifier (e.g SVM [25]) to train over subset of features and other classifier (e.g Naive Bayes) to test optimal set of features, but embedded feature selection approaches are bound to use the same classifier throughout the classification process [69].

On the other hand, filter based feature selection algorithms do not take into account the error value of a classifier, however, they rank the features and pick top $k$ features based on certain threshold [71]. In this way, a highly discrimi-native user specified subset of features is acquired by utilizing the statistics of data samples.

Wrapper, and embedded feature selection methods are computationally far more expensive as compared to filter based feature selection algorithms. While both former approaches assess the usefulness of features by cross validating classifier performance, latter approaches operates over the intrinsic properties (e.g relevance) of features computed through univariate statistics.

In our work, considering the efficiency of filter based feature selection algorithm, we have adapted ten most anticipated filter based feature selection algorithms. These algorithms are extensively being utilized for English text document classification such as Balanced Accuracy Measure (ACC2) [45], Nor-malized Difference Measure (NDM) [43], Max-Min Ratio (MMR) [21], Relative Discrimination Criterion (RDC) [46], Information Gain (IG) [47], Chi-Squared (CHISQ) [44], Odds Ratio (OR) [48], Bi-Normal Separation (BNS) [45], Gini Index (GINI) [49], Poisson Ratio (POISON [50]) [50]. Here, we only refer these feature selection algorithms, interested readers can explore these algorithms deeply by studying their respective papers.

## 3.4 Feature Representation

Diverse domains (e.g textual, non-textual) have different stacks of features, for example, if we want to classify iris data then the set of useful features would be sepal length, sepal width, petal length and petal width [72]. However, the set of textual features for certain domain is not fixed at all. Representation of features plays a vital role to raise the performance of diverse classification methodologies [29], [58], [59]. Machine learning methodologies utilize bag of words based feature representation approaches. Term frequency [73] is the simplest and widely used feature representation technique for various natural language processing tasks such as text classification and information retrieval [[74], [75], [76]]. Term frequency (TF) [73] of a term in a document is defined as the number of times a term occur in that document. One of the most significant problem of TF is that it does not capture the actual importance and usefulness of a term. This down-fall is well addressed by Term Frequency-Inverse Document Frequency (TF-IDF) [73] which is a modified version of term frequency [73] as it declines the weight specifically for the words which are commonly used and raises the weight for less commonly used words of underly corpus. It gives more importance to less frequent terms and vice versa. It is calculated by taking dot product of term frequency (TF) and inverse document frequency(IDF).

IDF assigns weights to all the terms on corpus level. According to IDF, a term is more important if it occurs in less documents. When IDF weighting scheme is used standalone, it can allocate same weights to many terms which have same $DF_t$ score. IDF is defined as follows:

$$IDF_t = \log \frac{N}{DF_t} + 1 \quad (1)$$

where $N$ is the total number of documents in the corpus and $DF_t$ is the document frequency of term $t$.

A higher TF-IDF score implies that the term is rare and vice versa. Its value for term $t$ in a document $d$ can be calculated as

$$TF - IDF_{t,d} = TF_{t,d} \cdot IDF_t \quad (2)$$

Thus, by using both TF and IDF, TF-IDF captures the actual importance of terms on both document and corpus level.

## 3.5 Classifiers

In order to assess the impact of filter based feature selection algorithms on the performance of trivial machine learning based Urdu text document classification methodologies, we utilize Support Vector Machine (SVM) [25], and Naive Bayes (NB) [26] classifiers. This is because, in state-of-the-art Urdu text document classification work, we have found that only these two classifiers mark promising performance [14],[15],[17],[16],[18],[19].

Naive Bayes [26] uses bayes theorem and probability theory in order to make predictions. Naive Bayes [26] classifiers are usually categorized as *Generative Classifiers* and are highly useful for applications like document classification [77], and email spam detection [78]. Whereas, SVM [79] classifier is categorized as *Discriminative Classifier* and mostly used for anomaly detection [80], and classification problems [81]. It is a non probabilistic linear classifier which plots each data sample as a coordinate point in multi dimensional space and finds an optimal hyper plane which eventually helps to differentiate the class boundaries effectively.

## 3.6 Adopted Deep Learning Methodologies For Urdu Text Document Classification

This section summarises state-of-the-art deep learning based methodologies adapted for the task of Urdu text document classification. In order to provide a birds eye view on adopted deep learning methodologies, generalized architecture is drawn in Figure 2

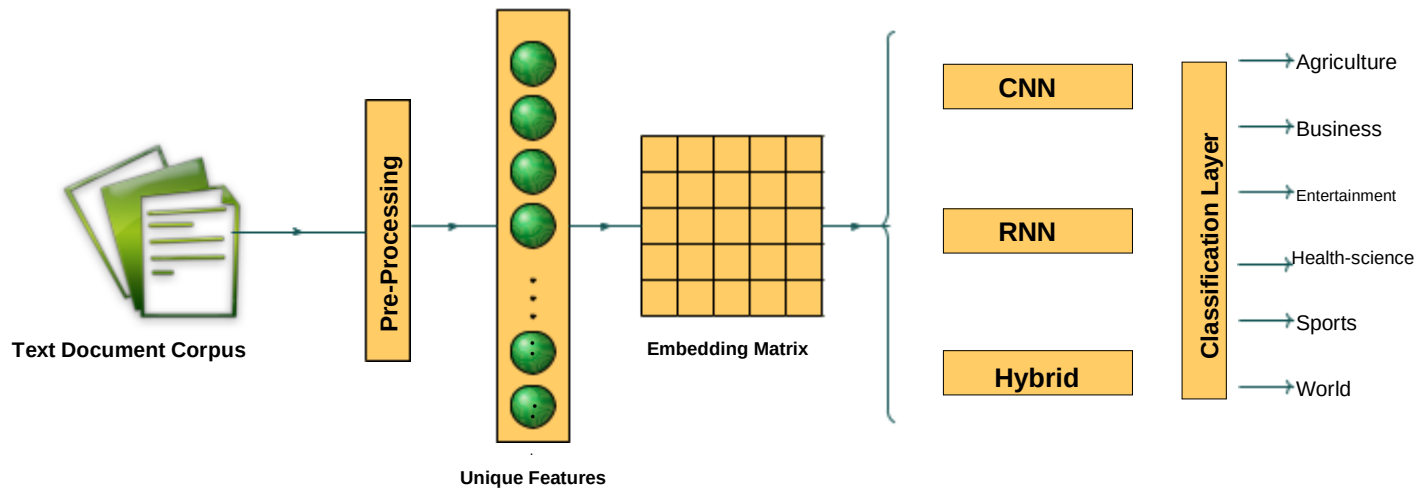

Figure 2: Generalized Methodology of Adopted Deep Learning Models

We adapt a multi-channel CNN model presented by Yoon Kim [82] for the task of sentiment, question, and sentence classification. In order to reap the benefits of distinct pre-trained word vectors, for the very first time, they made few channels dynamic, and others static throughout training in order to prevent overfitting. Several researchers (e.g Nabeel et al. [42]) utilized this model for English text document classification and achieved state-of-the-art performance. In our experimentation, we have fed FastText embeddings at one channel and pre-trained embeddings provided by Haider et al. [32] at second channel. At third channel, we have used randomly initialised word embeddings. In order to avoid overfitting, we keep the FastText embeddings static, and fine tuned other embeddings during training.

Embedding layer of this model is followed by 3 convolution layers with 128 filters of size 3, 4 and 5 respectively. After that, extracted features of all convolution layers are concatenated and fed to another convolution layer having 128 filters of size 5. After applying max-pooling of size 100, the extracted features are then passed to a flatten layer which flattens the features. These flattened features are then passed to a dense layer with 128 output units which are followed by a dropout layer of rate 0.5. Finally, a last dense layer acts as a classifier.

Another CNN based approach adapted for Urdu text document classification was presented by Nal Kalchbrenner et al. [83]. A distinct aspect of this model was the use of wide convolutions. The authors claimed that the words at edges of a document do not actively participate in convolution and get neglected especially when the filter size is large. An important term can occur anywhere in the document so by using wide convolution every term take equal part while convolving. Although, originally, authors did use any pretrained word embed-dings in the proposed CNN architecture, however, we have utilized pretrained word embeddings.

This model begins with an embedding layer followed by convolution layer with 64 filters of size 50. Top five features are extracted from the convolution layer by using a K-max-pooling layer of value 5. Zero padding is utilized to maintain the wide convolution structure. After that, there is another convolution layer with 64 filters of size 25. This layer is followed by a K-max-pooling layer of value 5. Finally, the extracted features are flattened and passed to a dense layer which classifies the documents.

Yin et al. [84] proposed a CNN model for the task of binary or multi-class sentiment analysis, and subjectivity based question classification for the English Language. The significance of the multi-channel input layer was deeply explored by the author by using five different pre-trained word vectors. This model has outperformed eighteen baseline machine and deep learning methodologies [84] for sentiment and question classification tasks. While adopting this model, we have utilized two embedding layers, two convolution layers along with wide convolutions.

The model starts with two embedding layers, each embedding layer is followed by two wide convolution layers with 128 filters of size 3 and 5 respectively. Each convolution layer is followed by a K-max-pooling layer of size 30. After that, both convolution layers are followed by two other convolution layers of same architecture except the value of $k$ which is 4 in K-max-pooling layers. All the features from all convolution layers are then concatenated and flattened by using a flatten layer. These flattened features are then passed to two dense layers from which the first dense layer has 128 output units and the last dense layer acts as a classifier.

Just like Yin et al. [84] CNN based approach, Zhang et al. [85] also proposed a CNN based approach for text classification. In proposed approach, they not only experimented with three different pre-trained neural word embeddings but also applied l2 norm regularization before and after concatenating all features of different channels. While adopting this model in our experimentation, three embedding layers, l2 norm regularization after features concatenation, and wide convolutions are utilized.

The model starts with three embedding layers and each embedding layer is then followed by two convolution layers. Both convolution layers have 16 filters of size 3 and 5 respectively which are followed by a global max-pooling layer. After that, features of all layers are concatenated and l2 norm regularization is applied using a dense layer with 128 output units. These features are then passed to a dense layer which acts as a classifier.

Dani Yogatama el al., [86] proposed an LSTM based neural network model for classifying news articles, questions, and sentiments. Two different versions of the model namely generative and discriminative LSTM model were proposed. Both models were the same except that the discriminative model tried to maximize the conditional probability while the generative model maximized the joint probability. We adopt discriminative version of the model. This model begins with an embedding layer and output of the previous layer is fed to an LSTM layer which has 32 units. The features extracted by LSTM are then flattened and passed to a dense layer for classification.

Another LSTM based model was proposed by Hamid Palangi et al., [87] to generate the sentence neural embeddings for raising the performance of document retrieval task. This model was not used for any sort of text classification but as its architecture is pretty similar to Yogatama el al., [86] proposed model that is why we have adopted this model for our experimentation. The output of the first embedding layer is fed to an LSTM layer which has 64 output units. The output of the LSTM layer is then flattened and feed into a dense layer that acts as a classifier.

As discussed before, both CNN and RNN have their own benefits and draw-backs [88]. In order to reap the benefits of both architectures CNN, and RNN, re-searchers proposed hybrid models [89],[90],[91],[88],[92] in which usually a CNN architecture is followed by RNN. CNN extracts global features [93], [94], [89], while RNN learns long term dependencies for the extracted features [95] [96] [97], [98], [99][100] [101] [102].

A hybrid model was presented by Siwei Lai et al., [103] for the task of text classification. The author claimed that RNN was a biased model in which later words were more dominant than earlier words. To tackle this problem a hybrid model was suggested that consists of bi-directional LSTM followed by a max-pooling layer. The bi-directional nature of the model reduces the words dom-inance whereas max-pooling layer captures more dicriminative features. This model has outperformed twelve machine and deep learning based models for the task of text classification.

The model begins with three embedding layers, first one is passed to forward LSTM layer and the second one is fed to backward LSTM layer. Both LSTM layers have 100 output units. The yielded features from both LSTMs are con-catenated along with third embedding layer and pass to a Dense layer which has 200 output units. Dense layer is followed by a max-pooling layer and the output of max-pooling layer is then passed to another dense layer which acts as a classifier.

Guibin Chen et al., [104] proposed another hybrid model that consists of CNN and LSTM and used for multi-label text classification. Pre-trained word embeddings were used to feed the CNN and then features were extracted to feed LSTM. The author claimed that the pre-trained word vectors contain the local features of each word whereas CNN captured the global features of the input document. Both local and global features were then used by LSTM to predict the sequence of labels. We have adopted this model for multi-class classification instead of multi-label classification.

The model starts with an embedding layer which is followed by five convo-lution layers with 128 filters of sizes 10, 20, 30, 40, and 50 respectively. Each convolution layer is followed by a max-pooling layer of the same filter size. The output features from all five max-pooling layers are concatenated and flattened using a flatten layer. These flattened features are then passed to a dense layer which has 128 output units. The output from the dense layer along with the output of the embedding layer is then passed to an LSTM layer. This LSTM layer is followed by another dense layer that acts as a classifier.

Another hybrid model based on CNN and LSTM was proposed by Chunting Zhou et al., [105] for sentiment analysis and question classification. CNN was used to capture the high level word features whereas LSTM extracted the long term dependencies. Different types of max-pooling layers were applied to the features extracted from CNN. However, the authors suggested that max-pooling layer must be avoided if the features needed to be passed to LSTM. Because LSTM was used for sequential input and a max-pooling layer would break the sequential architecture.

The output of the first embedding layer is passed to five convolution layers which have 64 filters of size 10, 20, 30, 40, and 50 respectively. The extracted features of these five convolution layers are then concatenated and fed to an LSTM layer which has 64 output units. This layer is followed by two dense layers from which the first dense layer has 128 units and the last dense layer eventually acts as a classifier.

The last chosen model in our research is also a hybrid model presented by Xingyou Wang et al. [106] for sentiment classification. The theory behind this model is the same as Chunting Zhou et al., [105] model except it used both LSTM and GRU along with max-pooling layers after CNN. Based on experimental results, authors claimed that both LSTM and GRU produced the same results, that is why we have adopted this model only with LSTM for our experimentation.

This model begins with an embedding layer followed by three convolution layers which have 64 filters of size 3, 4 and 5 respectively. Each convolution layer is followed by a max polling layer of same filter sizes. After that, all the output features of the max-pooling layers are concatenated and passed to an LSTM layer which has 64 units. The features yielded by LSTM layer is then passed to a dense layer which has 128 units. This layer is followed by another dense layer that finally acts as a classifier.

## 3.7 Transfer Learning Using BERT

This section discusses the fruitfulness of transfer learning using pretrained lan-guage model “BERT [38]” for the task of Urdu text document classification. Pre-training language model has proven extremely useful to learn generic language representations. In previous section, all discussed deep learning based classi-fication methodologies utilized pre-trained neural word embeddings including Word2vec [107], FastText [108], and Glove [109]

Traditional neural word embeddings are classified as static contextualized embeddings. These embeddings are prepared by training a model on a gigantic corpus in an unsupervised manner to acquire the syntactic and semantic prop-erties of the words up to certain extent. However these embeddings fail to grasp polysemy which is all about generating distinct embeddings for the same word on account of different contexts [33], [34], [35], [33]. For instance, consider two sentences like “Saim, I ’ll get late as I have to deposit some cash in Bank” and the other one is “My house is located in canal Bank”. In both sentences, word Bank has a different meaning. However, models build on top of neural word embeddings do not consider the context of words in which they appear, thus in both sentences the word “Bank” will get a same vector representation which is not correct.

These downfalls are resolved by pre-trained language models which learn the vector representation of words based on the context in which they appear and this is why embeddings of pre-trained language models such as Bidirectional

Encoder Representations from Transformers (BERT [38]) are categorized as dynamic contextualized embeddings. Dynamic contextualized embeddings capture word semantics in dissimilar contexts to tackle the problem of polysemous, and context dependent essence of words. In this way, language models such as BERT [38] manages to create different embeddings for the same word which appear in multiple contexts. Traditional language models are trained from left to right, thus they are framed to predict next word. Contrarily, there exist few approaches such as Universal Language Model Fine-Tuning (UMLFit) [37] and Embeddings for Language Models (ELMo) [36] based on Bi-LSTM. Bi-LSTM is trained from left to right in order to predict next word, and from right to left to predict previous word, however not both at the same time. Whereas, BERT [38] utilizes entire sentence to learn from all words located at different positions. It randomly masks the words in certain context before making prediction. In addition, it uses transformers which further make it accurate.

To summarize, due to masked language modelling, BERT [38] supersedes the performance of other language modelling approaches such as UMLFiT [37], and ELMO [36]. Moreover, training the transformed architecture bidirectionally in language modelling has proved extremely effective as it has deeper understanding of language context than uni-directional language models. Although BERT [38] has marked promising results in several natural language processing (NLP) tasks, nevertheless, there exists a limited research to optimize BERT [38] for the improvement of target NLP tasks. In this paper, we thoroughly investigate how to make the best use of BERT [38] for the task of text document classi-fication. We explore multifarious methods to fine-tune BERT [38] in order to maximize its performance for Urdu text document classification. We perform pre-processing in a same manner as discussed in detail in the section 3.2

## 3.8 Hybrid Methodology For Urdu Text Document Classification

This section explains the hybrid methodology for the task of Urdu text doc-ument classification. It is considered that deep learning based methodologies automate the process of feature engineering, however, recent research in com-puter vision [110], and natural language processing (NLP) [42] extrapolates that these methodologies also extract some irrelevant and redundant features too which eventually derail the performance of underlay methodologies. In NLP, to remove irrelevant and redundant features, we [42] proposed a hybrid methodol-ogy which harvested the benefits of both trivial machine learning based feature engineering, and deep learning based automated feature engineering. In pro-posed hybrid methodology, first, a vocabulary of discriminative features was developed by utilizing a filter based feature selection algorithm namely Normal-ized Difference Measure (NDM) [43] and then the constructed vocabulary was fed to the embedding layer of CNN. Hybrid methodology managed to produce the promising figures on two benchmark English datasets 20-Newsgroup [7], and

[7]http://archive.ics.uci.edu/ml/datasets/twenty+newsgroups

BBC [8], when compared against the performance figures of traditional machine, and deep learning methodology. To evaluate that the proposed hybrid approach is extremely versatile and its effectiveness is neither biased towards the size of training data nor towards specific language or deep learning model, we assess the integrity of hybrid methodology by performing experimentation on different datasets and language with a variety of deep learning models. We adopt 4 CNN, 2 RNN, and 4 Hybrid models (CNN+RNN) which were previously used for text document or sentence classification (discussed in section 3.6). Hybrid approach is evaluated on three Urdu datasets (CLE Urdu Digest $1000k$, CLE Urdu Digest $1M$, DSL Urdu News).

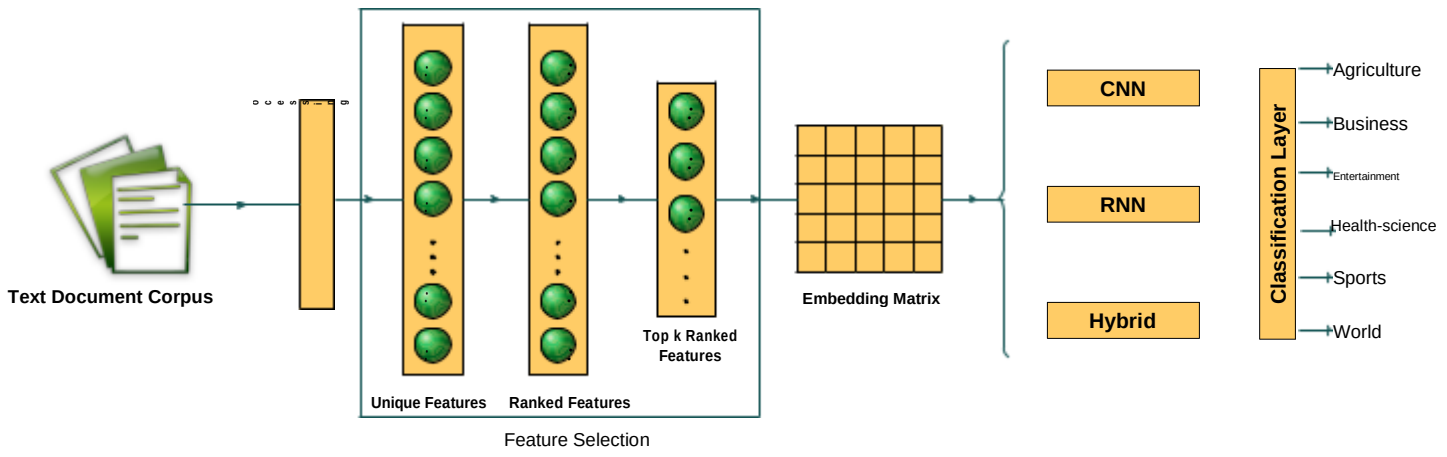

Figure 3: Machine And Deep Learning Based Hybrid Methodology [42]

We perform pre-processing in a same manner as discussed in detail in the section 3.2.

# 4 Datasets

To evaluate the integrity of all three methodologies based on machine learning, deep learning, and hybrid approach, we use two state-of-the-art closed source corpora CLE Urdu Digest $1000k$, CLE Urdu Digest $1M$, and one publicly available presented corpus namely DSL Urdu news. All textual documents of DSL Urdu news dataset are crawled from following web sites Daily Jang[9], Urdu Point[10], HmariWeb[11], BBC Urdu[12], and parsed through Beautiful Soup[13]. Table 2 illustrates the characteristics of newly developed corpus hav-ing $300K$ words, 4224 sentences, and a total 662 documents which belong to following six categories health-science, sports, business, agriculture, world and entertainment. Average length of a document is approximately 193 words in the developed corpus.

State-of-the-art corpora CLE Urdu Digest 1000K contain 270 news documents, and CLE Urdu Digest $1M$ contain 787 news documents belonging to 8 classes. Former one is a precise corpus and average length of a document is nearly 140 words, however, latter one is a large corpus with an average docu-ment length of 900 words. Statistics of both corpora with respect to each class are reported in the Tables 3, 4] respectively.

[8] http://mlg.ucd.ie/datasets/bbc.html
[9] https://jang.com.pk/
[10] https://www.urdupoint.com/
[11] http://hamariweb.com/
[12] https://www.bbc.com/urdu
[13] https://pypi.org/project/beautifulsoup4/

| Class | No. of documents | No. of sentences | No. of tokens | No. of tokens After lemmatization |
|---|---|---|---|---|
| Agriculture | 102 | 669 | 17967 | 9856 |
| Business | 120 | 672 | 20349 | 9967 |
| Entertainment | 101 | 685 | 19671 | 10915 |
| World | 111 | 631 | 18589 | 12812 |
| Health-sciences | 108 | 823 | 27409 | 12190 |
| Sports | 120 | 744 | 24212 | 9992 |

Table 2: DSL Urdu News Dataset Statistics

| Class | No. of documents | No. of sentences | No. of tokens | No. of tokens After lemmatization |
|---|---|---|---|---|
| Culture | 28 | 488 | 8767 | 8767 |
| Health | 29 | 608 | 9895 | 9895 |
| Letter | 35 | 777 | 11794 | 11794 |
| Interviews | 36 | 597 | 12129 | 12129 |
| Press | 29 | 466 | 10007 | 10007 |
| Religion | 29 | 620 | 9839 | 9839 |
| Science | 55 | 468 | 8700 | 8700 |
| Sports | 29 | 588 | 10030 | 10030 |

Table 3: CLE Urdu Digest $1000k$ dataset statistics before and after Lemmatization

| Class | No. of documents | No. of sentences | No. of tokens | No. of tokens After lemmatization |
|---|---|---|---|---|
| Culture | 133 | 8784 | 145228 | 145228 |
| Health | 153 | 11542 | 169549 | 169549 |
| Letter | 105 | 8565 | 115177 | 115177 |
| Interviews | 38 | 2481 | 41058 | 41058 |
| Press | 118 | 6106 | 125896 | 125896 |
| Religion | 100 | 6416 | 107071 | 107071 |
| Science | 109 | 6966 | 117344 | 117344 |
| Sports | 31 | 2051 | 33143 | 33143 |

Table 4: CLE Urdu Digest 1M dataset statistics before and after Lemmatization

# 5 Experimental Setup and Results

This section summarizes different APIs that are used to perform Urdu text document classification. It also discusses the results produced by methodolo-gies based on machine learning, deep learning, and hybrid approach on three datasets (DSL Urdu news, CLE Urdu Digest $1000k$, CLE Urdu Digest $1M$) used in our experimentation. In order to process Urdu text for the task of Urdu text document classification, we develop a rule base sentence splitter and tokenizer. To evaluate the integrity of machine learning based Urdu text document classification methodology, all three datasets are splitted into train and test sets containing 70%, and 30% documents from each class respectively. The parameters of Naive Bayes [26] classifier are alpha=1.0, fit_prior=True, class_prior=None, and SVM [25] classifier is used with linear kernel and balanced class weight.

On the other hand, in order to evaluate the performance of adopted deep learning methodologies and to perform a fair comparison with machine learning based approaches for all three datasets, we use 30% data for test set and remain-ing 70% data is further splitted into train and validation sets having 60% and 10% data respectively. We use Keras API to implement the methodologies of ten adopted neural network based models. Pre-trained Urdu word embeddings provided by Haider et. al [32], and FastText [14] are used to feed all embedding layers except the second layer in Yin et al., [84] model and both second and third layers in Zhang et al., [85] model which are randomly initialized. To eval-uate and compare the performance of filter based feature selection algorithms, first we rank the features of training corpus against all classes. Then, at differ-ent predefined test points, we take top $k$ features from all classes and feed these features to two different classifiers SVM [25], and Naive Bayes [26]. For adopted deep learning based Urdu text document classification methodologies, we per-form experimentation in two different ways. In first case, after pre-processing, we select entire set of unique terms of each corpus and fed to the embedding layer of all adopted models (discuss with detail in section 3.6). Whereas, in second case, we select 1000 most frequent terms for DSL Urdu News, and CLE Urdu Digest $1000k$ datasets, and $10,000$ most frequent terms for CLE Urdu Digest $1M$ dataset.

Likewise, to evaluate the performance of hybrid approach which reaps the benefits of both machine and deep learning based feature engineering, as similar to machine learning based classification, for each dataset, we first rank the features of training corpus using NDM [43] feature selection algorithm. Then, top $k$ features of each class are fed to 10 different deep learning models. Rather then performing extensive experimentation with all feature selection algorithms once again, considering the promising performance produced by NDM [43] with all machine learning based methodologies, we only explore the impact of NDM [43] feature selection algorithm for 10 different deep learning based classification methodologies.

To assess the effectiveness of transfer learning using BERT [38], we fine-tune multilingual cased language model (BERT-Base [38]) having 12-layers, 12, heads, 768 hidden units, 110M parameters and pre-trained on 104 languages. We utilize multilingual cased model as it resolves normalization problems in several languages. We fine-tune multilingual model with the buffer size of 400, sequence length of 512, batch size of 16, and learning rate of 1e-5 for 50 epochs.

As two close source experimental datasets (CLE Urdu Digest $1000k$, $1M$) are highly unbalanced, thus instead of using accuracy, or an other evaluation mea-sure, we have performed evaluation using F1 measure as it is widely considered more appropriate evaluation measure for un-balanced datasets.

[14] https://github.com/facebookresearch/fastText/blob/master/pretrained-vectors.md

## 6 Conclusion

This paper may be considered a milestone towards Urdu text document classification as it presents a new publicly available dataset (DSL Urdu News), introduces 10 filter based feature selection algorithms in state-of-the-art ma-chine learning based Urdu text document classification methodologies, adopts 10 state-of-the-art deep learning methodologies, asseses the effectiveness of trans-fer learning using BERT, and evaluates the integrity of a hybrid methodology which harvests the benefits of both machine learning based feature engineering, and deep learning based automated feature engineering. Experimental results show that in machine leaning based Urdu text document classification methodol-ogy, SVM classifier outperforms Naive Bayes as all feature selection algorithms produce better performance for two datasets (CLE Urdu Digest $1000k$, $1M$) with SVM classifier. NDM and CHISQ reveal the promising performance with both classifiers. Amongst all, GINI shows the worst performance with both classifiers. Furthermore, adopted deep learning methodologies fail to mark a promising performance with trivial automated feature engineering. Although, using a vocabulary of most frequent features raises the performance of adopted deep learning methodologies, however it fails to obliterate the promising per-formance figures of hybrid approach. The hybrid methodology has proved ex-tremely versatile and effective with different languages. It substantially outper-forms adopted deep learning based methodologies and almost equalize the top performance of machine learning methodologies across two datasets (DSL Urdu News, CLE Urdu Digest $1M$). Similarly, BERT almost mimics the performance of hybrid methodology on account of those datasets where the average docu-ment length does not exceed 512 tokens. However for datasets where average document length exceeds from 512 tokens, hybrid methodology performs bet-ter than BERT. Contrarily, for all three datasets, hybrid methodology fails to outshine the peak performance figures produced by machine learning method-ology due to the small size of experimental datasets. To illustrate the point, consider the class Interviews of CLE Urdu Digest 1M which has only 38 doc-uments, so in this scenario, deep learning based hybrid methodology only uses 22 documents for training which are not good enough at all. A compelling future line of this work would be the development of a robust neural feature selection algorithm which can assists the models to automatically select highly discriminative features from each class. In addition, investigating the impact of ensembling feature selection algorithms over the performance of Urdu text document classification will also be interesting.